# MinkSORT: A 3D deep feature extractor using sparse convolutions to improve 3D multi-object tracking in greenhouse tomato plants

David Rapado-Rincon, Eldert J. van Henten, Gert Kootstra

**Abstract**

The agro-food industry is turning to robots to address the challenge of labour shortage. However, agro-food environments pose difficulties for robots due to high variation and occlusions. In the presence of these challenges, accurate world models, with information about object location, shape, and properties, are crucial for robots to perform tasks accurately. Building such models is challenging due to the complex and unique nature of agro-food environments, and errors in the model can lead to task execution issues. In this paper, MinkSORT, a novel method for generating tracking features using a 3D sparse convolutional network in a deepSORT-like approach, is proposed to improve the accuracy of world models in agro-food environments. MinkSORT was evaluated using real-world data collected in a tomato greenhouse, where it significantly improved the performance of a baseline model that tracks tomato positions in 3D using a Kalman filter and Mahalanobis distance. MinkSORT improved the HOTA from 42.8% to 44.77%, the association accuracy from 32.55% to 35.55%, and the MOTA from 57.63% to 58.81%. Different contrastive loss functions for training MinkSORT were also evaluated, and it was demonstrated that it leads to improved performance in terms of three separate precision and recall detection outcomes. The proposed method improves world model accuracy, enabling robots to perform tasks such as harvesting and plant maintenance with greater efficiency and accuracy, which is essential for meeting the growing demand for food in a sustainable manner.

**Keywords**: world modelling, deepsort, robotics in agriculture, deep learning, 3D deep learning

**List of Nomenclature**

| | |
|---|---|
| HOTA | Higher Order Tracking Accuracy |
| IDSW | ID Switches |
| LiDAR | Light Detection and Ranging |
| MHT | Multiple Hypothesis Tracking |
| MOT | Multi-Object Tracking |
| MOTA | Multi-Object Tracking Accuracy |
| RGB | Red Green Blue |
| SORT | Simple Online Realtime Tracking |
| XYZRGB | Channels X, Y, Z, and Red, Green and Blue on a point cloud |

# 1. Introduction

The global demand for food has been steadily increasing, as the world's population continues to grow and become more affluent. At the same time, the agro-food industry is facing a shortage of labour, particularly in labour-intensive production systems such as greenhouses. This shortage is caused by a variety of factors, including declining rural populations, increased competition for workers from other industries, and aging farm populations (Ince Yenilmez, 2015). In order to meet the growing demand for food, while also addressing the labour shortage, the agro-food industry is turning to technology, particularly robots, as a solution (Bogue, 2016). However, robots face many challenges in agro-food environments, such as the high degree of variation and occlusions, making it difficult for robots to accurately perceive and interact with these objects (Bac et al., 2014; Kootstra et al., 2020).

A key component for robots to succeed in these challenging environments is having a good and accurate representation of the environment, often referred to as a world model (Crowley, 1985; Elfring et al., 2013). In robotics, a world model is a representation of the physical environment that a robot is operating in, which can include information about the location, shape, and properties of objects in the environment. Robots require an accurate and up-to-date world model in order to function effectively. Without this information, robots may struggle to perform tasks accurately and may even cause harm to the environment or themselves. Conventional agro-food robotic systems have used simple world models based on a single sample of sensor data, like an image. From it they attempt an action, and then discard the previously obtained information. These traditional strategies suffer a substantial decrease in performance in highly occluded agro-food environments (Arad et al., 2020). More advanced world models that keep a memory of past information (Rapado-Rincón et al., 2023) can benefit from multi-view and active perception approaches (Burusa et al., 2022) that improve the detection and localization performance in occluded agro-food environments. This highlights the importance of developing robust world models for robots operating in challenging environments such as agro-food systems. In the agro-food context, an accurate world model could include information about the location of plants, the location of the fruits, the ripeness stage of the fruits, and other conditions of the plants, such as their growth stage and health; and maintain and update all this information over time. This would enable the robot to perform tasks such as harvesting and plant maintenance with greater accuracy and efficiency. For instance, it allows a robot to reason about a complex environment and choose the best path to a partially occluded fruit, while conventional approaches might fail in detecting the occluded fruit at all.

A common approach to build world models containing all relevant objects in the environment is the use of a multi-object tracking (MOT) algorithm, which associates noisy and uncertain measurements from the robot's sensors with the corresponding objects (Elfring et al., 2013; Persson et al., 2020; Wong et al., 2015). MOT has been used in agro-food environments for tasks such as monitoring and fruit counting (Halstead et al., 2018, 2021; Kirk et al., 2021; Villacrés et al., 2023). In previous work (Rapado-Rincón et al., 2023), it was demonstrated how occlusions can greatly reduce the performance of a tracking algorithm used to build a representation of greenhouse tomato plants, and how more complex and powerful tracking algorithms might be needed. Wojke et al. (2017) developed deepSORT, a method to improve the data association performance of simple tracking algorithms based on the Kalman filter and the Hungarian algorithm, like SORT (Bewley et al., 2016), in the presence of occlusions. DeepSORT uses a small convolutional neural network to extract features of all detected objects. These features are then combined with positional information of the objects in the images to improve the tracking performance over occlusions. DeepSORT has been successfully adapted for use in agro-food environments (Kirk et al., 2021; Villacrés et al., 2023). (Kirk et al., 2021) developed a feature extractor network to track strawberries through different growth stages. To achieve this, the standard DeepSORT network was extended with a classification head that was trained to differentiate between different fruit IDs and ripeness stages simultaneously. The authors demonstrated that adding the classification head improved the data association capabilities of the feature extractor network. In Villacrés et al. (2023), different two-step tracking methods were compared for tracking apples in orchards. The results showed that when using a deep learning object detector, DeepSORT outperformed algorithms like standard SORT and multiple hypothesis tracking (MHT).

Most of the MOT tracking algorithms used in agro-food environments are based on 2D imaging systems, resulting in a representation that lacks the 3D geometric information that a robotic system may need to perform tasks such as harvesting or plant maintenance. Furthermore, existing MOT algorithms focus on video sequences with high frame rates and low frame-to-frame distance, which tends to simplify tracking. However, in robotic applications, differences between frames can be large due to the 3D nature of robot' motions and environments. This results in sequences with more persistent and complex occlusions, and sudden changes of viewpoints and perspectives. Tracking in 3D space adds an extra dimension that can help differentiate between objects, therefore, improving the tracking accuracy and robustness to the presence of occlusions and changes on camera perspectives. However, to the best of our knowledge, there are no applications of DeepSORT-like methods using 3D data such as point clouds. Therefore, in this work, MinkSORT, a DeepSORT-like novel method to generate features from point clouds using a Minkowski network with 3D sparse convolutions that can be used for tracking objects in 3D (Choy et al., 2019) is presented. This method has the potential to further improve the tracking performance of DeepSORT by 3D data to generate tracking features. The contributions of this work are as follows:

• A novel 3D sparse convolutional network, MinkSORT, that generates object tracking features from pointclouds of objects for a DeepSORT-like algorithm. It enables the use of 3D sensing data through data-driven deep learning algorithms to improve tracking performance.

• A comparison of the tracking performance between MinkSORT and a baseline without deep-learning-based object features (Rapado-Rincón et al., 2023). The comparison was performed using challenging real-world data collected in a tomato greenhouse.

• A comparison of the effects of different contrastive loss functions for the training of the feature extractor network.

• An evaluation of the tracking performance of MinkSORT when object detectors with different precision and recall performances are used. Given the known sensitivity of DeepSORT-like algorithms to the object detection performance, this experiment shows how sensitive MinkSORT is to changes in the object detector.

## 2. Materials and Methods

A scenario is assumed where a robot arm is tasked with monitoring, maintenance, or harvesting operations in a tomato greenhouse. It is assumed that the robot is positioned approximately in front of the target plant, and that multiple viewpoints are collected by it in order to remove occlusions and build an accurate world model of the target plant. In this section, how data was collected and annotated is explained, how the world model is created and maintained is detailed, and which experiments were performed to evaluate the accuracy of the proposed algorithm is described.

### 2.1. Data collection and annotation

MinkSORT has two independent deep neural networks. First, objects are detected using

Mask RCNN, an instance segmentation deep neural network (He et al., 2017). Second, a set of features are generated for each object using a feature extraction deep neural network. The networks were trained using two different datasets collected from greenhouses at Wageningen University & Research:

- Tomato instance segmentation. A dataset containing 1204 colour images of tomato plants from several varieties. Part of this dataset was collected for the work of Afonso et al. (2019) and for our previous work (Rapado-Rincón et al., 2023). The images were collected using different Intel Realsense cameras, and were labelled for instance segmentation.

- Tomato 3D tracking. This dataset was collected during our previous work (Rapado-Rincón et al., 2023) using an ABB IRB1200 robotic arm with an Intel Realsense L515 LiDAR camera attached to the end-effector. It contains both colour and 3D information of 100 viewpoints of seven different plants, totalling 700 images. The viewpoints were collected using a semi-cylindrical path as show in Figure 1. Examples of some of the viewpoints can be found in Figure 2. Individual tomatoes were labelled with bounding box and ID in the colour image. This dataset was split in two subsets: training set, with two plants and 1097 tomato instances, and test set with five plants and 3137 tomato instances. The subset with two plants was used to train MinkSORT, while the subset with five plants was used for evaluation. A larger test set was selected to better represent the challenge of variation in agri-food environments. By evaluating more plants, the performance of MinkSORT under different plants, trusses, tomatoes and occlusive leaves could be studied.

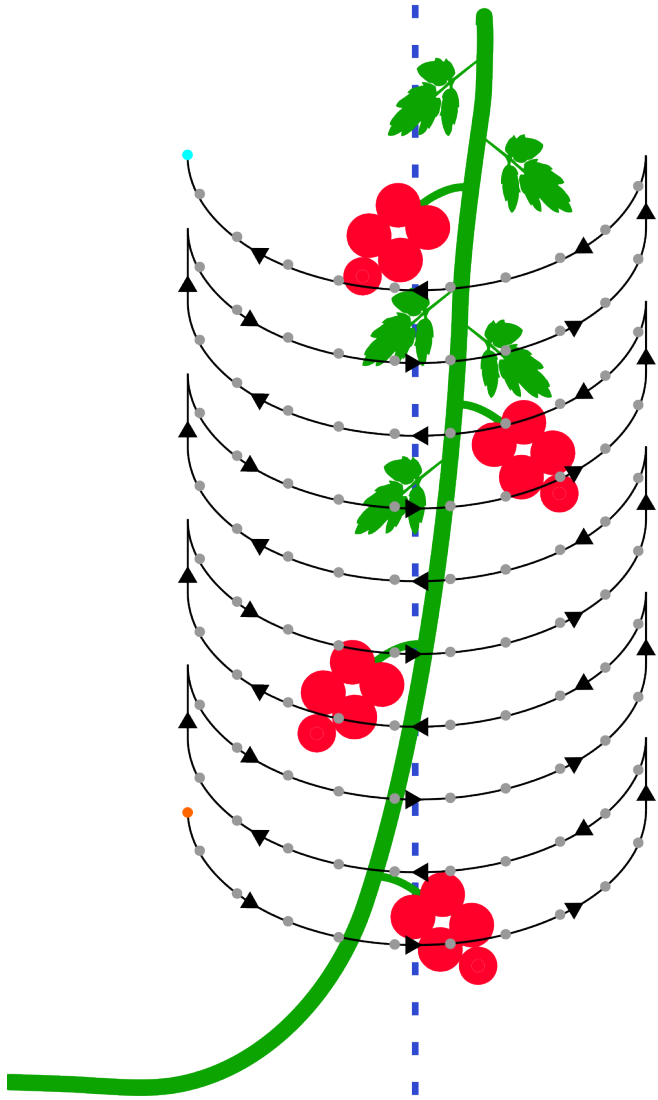

Figure 1. An example of the semi-cylindrical path used for collecting viewpoints for the tomato 3D tracking dataset is shown. The semi-cylinder was constructed using ten different heights, each with ten equally spaced viewpoints, represented by the grey dots. At each viewpoint, the camera was directed towards the centre of the semicircle of that height level, as represented by the blue dashed line.

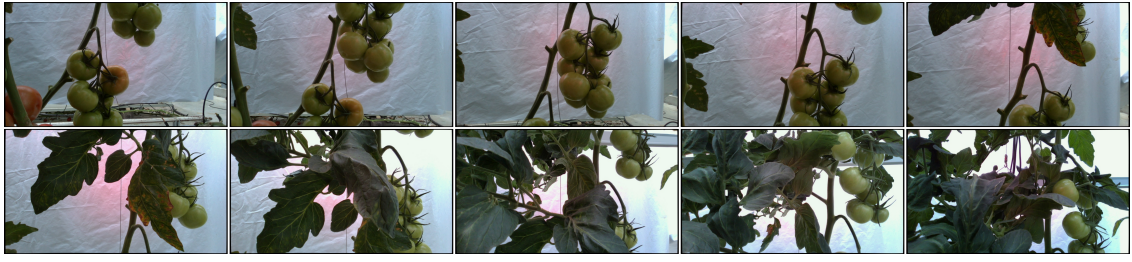

*Figure 2. Examples of viewpoints of a plant at each one of the tenth heights. The upper part of the plant illustrates a more complex scene due to occlusions by leaves.*

## 2.2. World Model

Figure 3 shows how the world model is maintained and updated over time using MinkSORT. At every frame, detections are generated using a tomato detector and a feature extractor. The tomato detection uses a combination of a deep-learning-based instance segmentation algorithm (He et al., 2017) and pointcloud processing methods to generate a list of tomato detections with their corresponding 3D position. The feature extractor is a deep neural network that given the pointcloud and colour information of each tomato, generates a feature vector. The detections are passed on to the data association algorithm, which associates existing tracks with the detections. Then these associations are used to update the world model by updating the objects. A prediction is made to project the tracks into the next time step, which is then used in the next cycle by the data-association algorithm together with detections from the next time step.

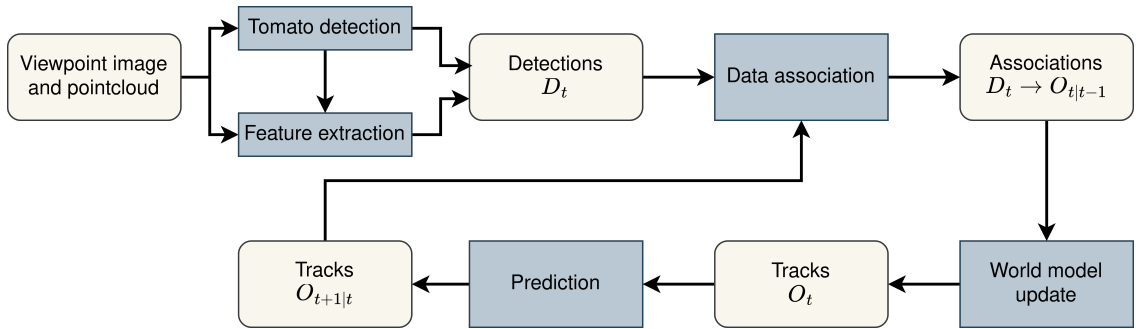

*Figure 3. Process used to create and update the robot world model by using a MOT algorithm, where the detections used for data association are generated by a two-step approach consisting of an object detector and an object feature extractor.*

The representation, or world model, of the robot's environment consists of a set of $M$ object representations $O_t = \{o_t^1, o_t^2, \ldots, o_t^M\}$ where $o_t^i = (p_t^i, F_t^i, b_t^i)$ corresponds to a specific object with position $p_t^i$, features list $F_t^i$, and bounding box $b_t^i$ at time $t$. From a tracking perspective, objects in the world model are also referred to as tracks. The position is represented as a Multivariate Gaussian distribution, $p_t^i = \{\mu_t^i, \Sigma_t^i\}$, where the mean, $\mu_t^i \in \mathbb{R}^3$, represents the most likely position of an object; and the covariance, $\Sigma_t^i \in \mathbb{R}^{3\times3}$, the uncertainty on the position. The object position is represented in the world model coordinate system, which corresponded to the robot coordinate system. $F_t^i$ corresponds to the list $\{f_1^i, f_2^i, \ldots, f_k^i\}$ that contains all the $k$ features associated to track $i$. The bounding box $b_t^i \in \mathbb{R}^4$ of an object was represented as the 2D bounding box of the last detection $d_t^j$ associated with that object in image coordinates.

### 2.2.1. Tomato detection

The detection system is based on our previous work (Rapado-Rincón et al., 2023). At every time step, the colour image and pointcloud are used to detect tomatoes. First, the colour image is passed into a Mask R-CNN algorithm that was trained to perform instance segmentation using the tomato instance segmentation dataset. The network generates tomato detections in the image with their corresponding object confidence, mask, and bounding box. The pointcloud is then filtered using the tomato masks to generate individual pointclouds per tomato. These are used in a sphere fitting algorithm to estimate the centre of each tomato. The output of the tomato detection is therefore a set of $N$ object detections at time $t$, $D_t = \{d_t^1, d_t^2, \ldots, d_t^N\}$ where $d_t^j = (p_t^j, b_t^j, m_t^j, f_t^j)$ corresponds to the detection $j$ at time $t$, including the 3D position of the detection with respect to the robot coordinate system $p_t^j$, bounding box $b_t^j$, mask $m_t^j$, and feature vector $f_t^j$. The position is defined as a Multivariate Gaussian distribution, $p_t^j = \{\mu_t^j, \Sigma_t^j\}$, with mean $\mu_t^j \in \mathbb{R}^3$ corresponding to the centre of each tomato, and covariance $\Sigma_t^j \in \mathbb{R}^{3\times3}$, which was estimated empirically. The feature vector $f_t^j$ is generated using the feature extraction network, which is described in the next section.

### 2.2.2. Feature extraction

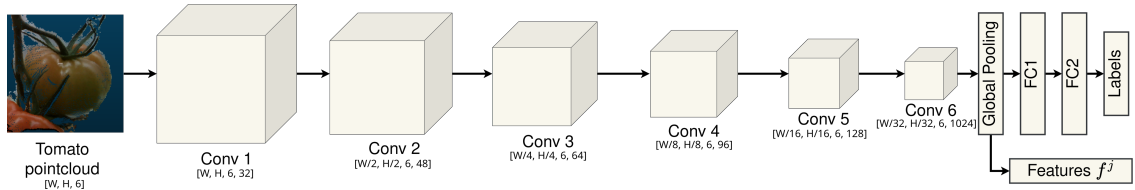

*Figure 4. Feature extraction network. It consists of six sparse convolutional layers, followed by a global pooling layer, and two fully connected layers. At training time, the output of the last fully connected layer, the class labels, together with the features generated after the global pooling, are used to train the network. At inference time, only the features are used.*

Inspired by the deep-learning-based feature extractors that have shown great performance in 2D MOT tasks (Wojke et al., 2017), a network that uses 3D data to generate tracking features of tomatoes was designed. As shown in Figure 4, MinkSORT is a Minkowski network with six sparse convolutional layers (Choy et al., 2019), a pooling layer, and two fully connected layers. In contrast to standard convolutions present in convolutional neural networks, sparse convolutions have been a popular method to build networks suitable for 3D data as the convolution kernel is only applied at occupied voxels. This is optimal for voxel representations of 3D data, as empty spaces and missing data are common problems. Sparse convolutions can therefore skip missing data which reduces computation and memory requirements. The network takes as input a six-channel pointcloud (XYZRGB), which corresponds to the pointcloud of a tomato detection. Similarly to DeepSORT, MinkSORT is trained as a supervised classification task, where the classes correspond to all the 58 different tomato IDs labelled in the dataset with two plants and 1097 individual tomato instances. This dataset was split in train and validation. At inference time, the fully connected layers are discarded as classification is not needed. Consequently, the output of the network is the feature vector $f_t^j$ of detection $j$ at time $t$. The progress of the performed trainings can be found in Figure 5.

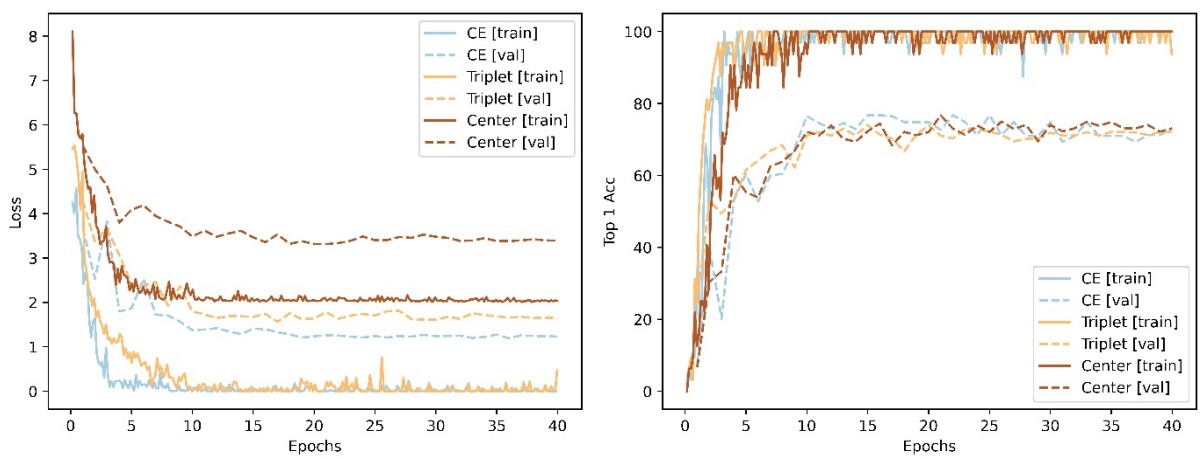

*Figure 5. Training and validation progress values. Left image shows the training and validation loss, while the right image shows the training and validation top 1 accuracy.*

During the experiments, several loss functions were evaluated: cross-entropy, centre loss (Wen et al., 2016), and triplet loss (Balntas et al., 2016). Centre loss and triplet loss are losses commonly used in tracking since they force the network to learn features that are spatially further apart between different classes. Centre and triplet loss are contrastive loss functions for tasks like image recognition or embedding learning, where the goal is to maximise the distinction between different classes in the learned feature space. They are often used in feature extraction networks for tracking as they maximise the distance between the feature vectors of two different object IDs. Centre loss tries to minimise intra-class differences by pulling class instances towards their respective class centres in the feature space. Triplet loss uses triplets of data: an anchor, a positive of the same class, and a negative of a different class; and ensures the anchor is closer to the positive than the negative in the feature space.

**2.2.3. Data association**

For every frame, detected objects $D_t$ are assigned to the existing tracks using their predicted state from the previous time step $O_{t|t-1}$ as $D_t \rightarrow O_{t|t-1}$. This association is computed using the Hungarian algorithm over the cost matrix $C(i,j)$ where each cell $(i,j)$ contains a cost value based on the position and features similarities between track $i$ and detection $j$. The cost matrix is computed as the weighted sum of the position cost matrix, $C_{pos}(i,j)$, and the features cost matrix, $C_{feat}(i,j)$, as

$$C(i,j) = (1 - \lambda)C_{pos}(i,j) + \lambda C_{feat}(i,j) \qquad (1)$$

where $\lambda$ is the parameter that controls the weighted sum, and $C_{pos}(i,j)$ is the squared Mahalanobis distance between the positions of track $i$ and detection $j$, calculated as

$$C_{pos}(i,j) = (\mu_t^j - \mu_t^i){\Sigma_t^i}^{-1}(\mu_t^j - \mu_t^i)^T \qquad (2)$$

and $C_{feat}(i,j)$ is the smallest cosine distance between the list features of track $i$, $F^i$, and the feature vector of detection $j$, $f^j$, calculated as

$$C_{feat}(i,j) = min\{1 - {f^j}^T f_k^i | f_k^i \in F^i\} \qquad (3)$$

Unlikely associations were discarded by setting a threshold to both position and feature cost matrices. Discarded associations $(i,j)$ are not considered in the Hungarian algorithm. For the position cost matrix, the threshold was set to 7.82, a value which corresponds to the 0.95 quantile of the chi-square distribution with 3 degrees of freedom as suggested by Wojke et al. (2017). Associations whose $C_{pos}(i,j)$ was larger than this threshold were discarded. The threshold for the feature matrix was computed using the validation subset derived from the set of two plants employed to train the feature extractor. The minimum cosine distance between all occurrences of all IDs in that validation set was calculated, and the 0.95 quantile of the resulting distribution was used as threshold.

**2.2.4. World model update and prediction**

After associating a track $o_t^i$ with a detection $d_t^j$, the attributes of the detection were utilised to update the object attributes. The feature vector $f^j$ of detection $j$ is added to the list of features $F_t^i$ of track $i$. The bounding box $b_t^i$ is replaced with the box of the associated detection, $b^j$. The object position $p_t^i$ is updated by using the standard Kalman filter update step, incorporating both the detection position $p_t^j$ and the predicted position of the object from the previous time step $p_{t|t-1}^i$.

When a detection could not be associated with any existing objects, a new object was initialised in the world model. The new object $o^i$ was initialised from the detection $d^j$. Since objects were assumed to not disappear from the robot environment, objects in the world model were never removed.

In the prediction step, objects in the world model were projected forward to the next time step. During this step, the object features $F_t^i$ and bounding box $b_t^i$ remain unchanged, and the object position $p_{t+1|t}^i$ was predicted using the standard Kalman Filter prediction step where the object is considered static. Therefore, the state of the Kalman Filter can be defined as $[x \quad y \quad z]$ Cartesian coordinates of the object with respect to the robot coordinate system.

**2.3. Experiments and evaluation**

The experiments were performed on the dataset of five plants, each with 100 viewpoints, explained in section 2.1. The High Order Tracking Accuracy (HOTA) (Luiten et al., 2021) and the Multi-Object Tracking Accuracy (MOTA) (Bernardin & Stiefelhagen, 2008), as well as their sub-metrics, were used to assess the performance of MinkSORT. HOTA considers both detection and association accuracy of objects over time by considering all frames at once. In HOTA, detection performance is measured using the sub-metrics Detection Recall (DetRe), Detection Precision (DetPr) and Detection Accuracy (DetA); while the association accuracy is measured using the Association Accuracy (AssA). MOTA measures the overall tracking accuracy, considering false positives (erroneously reported detections), misses (missed actual detections), and identity switches (IDSW), all of which reduce the MOTA score.

Two different experiments were performed. In the first experiment, to assess the impact of using contrastive losses, the performance of MinkSORT when different losses are used to train the deep-learning-based feature extractor was compared. In the second experiment, to analyse the effect of different detection performance on MinkSORT, three confidence thresholds for the Mask RCNN were used: 0.1, 0.5, and 0.9. These thresholds resulted in varying precision-recall values for the detection algorithm. For this experiment, only the feature extractor trained with cross-entropy loss was used.

Both experiments share the following methodology:

- To investigate the added benefit of using deep-learning-based features to track objects in 3D, MinkSORT was compared against a baseline algorithm that does not use deep-learning-based features. The baseline corresponds to the 3D tracking algorithm developed in our previous work (Rapado-Rincón et al., 2023), and it is referred to as 3D-SORT.

- To evaluate MinkSORT's performance under different average distances between viewpoints, two approaches to feed the sequence into it were employed:

    - Sequential order: The frames were processed in the same order as they were recorded in the semi-cylindrical path (see Figure 1).

    - Random order: The frames were processed in a random order. This increases the average distance between frames, as well as revisiting areas of the plant.

- Five sets of 80 out of the 100 viewpoints were selected, allowing us to perform statistical analysis on the performance results. A t-test was then used to assess whether the tested configurations of MinkSORT significantly improved or decreased the performance compared to the baseline.

## 3. Results

Table 1 shows how MinkSORT with a feature extractor trained using cross entropy significantly improves HOTA and MOTA over the baseline 3D-SORT for both sequential and random order frames. This shows that the feature extractor algorithm of MinkSORT is able to encode useful information that helps differentiate between different objects in a tracking task. When the order of the sequence is random, it can be seen how MinkSORT outperforms 3D-SORT independently of the loss function used during training. Furthermore, the difference between MinkSORT and the 3D-SORT is larger when a random sequence order is used. This suggests that MinkSORT is able to create a set of features that are more independent to the change of perspective than the estimated centre of the object that is generated from a partial pointcloud of the tomatoes.

Table 1. Tracking performance results obtained for MinkSORT when trained using three different loss functions, and evaluated over sequential and random sequences. A detection confidence threshold of 0.5 was used. Values with *, ** or *** mean that they significantly perform different than the baseline 3D-SORT with a P-value of 0.05, 0.01 and 0.001 respectively.

| Order | Model | HOTA↑ | DetRe↑ | DetPr↑ | AssA↑ | MOTA↑ | IDSW↓ |
|---|---|---|---|---|---|---|---|
| Sequential | 3D-SORT | 42.8 | 60.83 | 82.33 | 32.55 | 57.63 | 172.8 |
| | MinkSORT (CE) | 44.77** | 61 | 82.56 | 35.55** | 58.81* | 143.0** |
| | MinkSORT (CE + Center) | 41.31* | 61.03 | 82.6 | 30.35* | 49.2*** | 386.6*** |
| | MinkSORT (CE + Triplet) | 43.54 | 60.95 | 82.49 | 33.72 | 57.74 | 170 |
| Random | 3D-SORT | 24.79 | 60.86 | 82.36 | 11.01 | 29.84 | 876.8 |
| | MinkSORT (CE) | 29.6*** | 61 | 82.56 | 15.63*** | 31.01 | 847.4 |
| | MinkSORT (CE + Center) | 32.51*** | 61.05 | 82.63 | 18.89*** | 27.12** | 946.0** |
| | MinkSORT (CE + Triplet) | 31.16*** | 61 | 82.56 | 17.3*** | 33.48** | 785.0** |

Even though HOTA and MOTA significantly improve with MinkSORT, DetRe and DetPr do not show any significant changes between 3D-SORT and MinkSORT. This is expected, as they all share the same object detection algorithm, Mask RCNN. The main improvement of the proposed algorithm lays on the number of IDSW and the AssA. This is expected, as the goal of MinkSORT is to help differentiating between different objects hence increasing the data association performance.

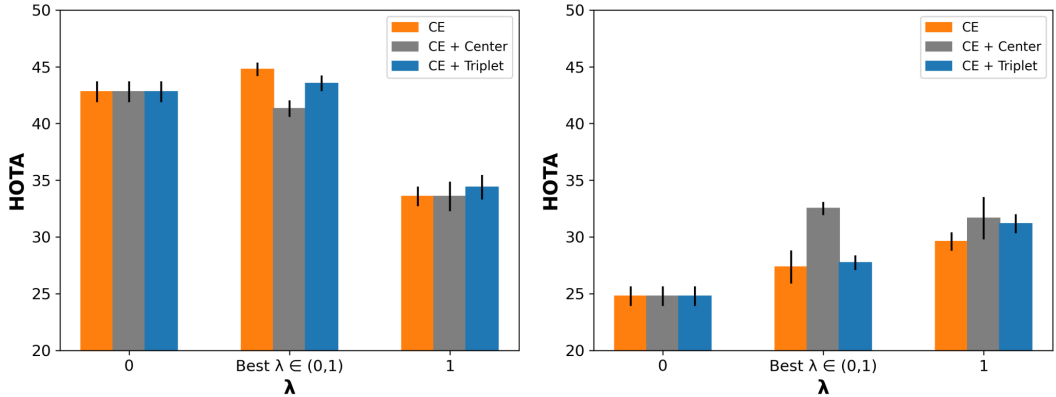

Figure 6. Bar plots containing the HOTA over different λ values and loss functions. The left plot shows the results when frames are processed in sequential order, while the right plot shows the equivalent when frames are processed in random order.

Figure 6 shows the performance of MinkSORT with feature extractors trained with different loss functions over three different $\lambda$ values in equation 1. A $\lambda$ of 0 corresponds to the baseline model, 3D-SORT, where the features generated by the feature extractor network are not used, and a $\lambda$ of 1 means that only features extracted from the network are used for tracking. Every other value in between means that both the position of the object and the deep

21

learning features are used for tracking. Since the value of $\lambda$ lays in continuous space, multiple values were sampled and selected the value with the highest HOTA performance for each loss configuration. Additionally, it can be observed that when $\lambda$ is set to 1, the performance of the tracking algorithm is very similar between sequential and random sequence orders. As discussed before, this suggests that the features generated by the feature extraction network are more invariant to the change in camera pose than the centre position of the objects. This is expected since the network is trained using random viewpoints of objects, without any sequence order at all.

Table 2 shows that, independently of the detection confidence threshold used, MOTA significantly improves, and the number of ID switches is significantly reduced with MinkSORT. HOTA and association accuracy also increase with all thresholds, but only with a threshold of 0.5 the improvement is significant. This can be due to the limited size of the evaluation dataset, which only includes five plants. These results show how MinkSORT can improve over 3D-SORT independently of the precision and recall performance of the detection algorithm.

Table 2. Tracking performance results obtained for MinkSORT over different detection confidence thresholds. The feature extractor used in the "deep" versions was trained using the cross-entropy loss. Values with *, ** or *** mean that they significantly perform different than the baseline 3D-SORT with a P-value of 0.05, 0.01 and 0.001 respectively.

| Threshold | Model | HOTA↑ | DetRe↑ | DetPr↑ | AssA↑ | MOTA↑ | IDSW↓ |
|---|---|---|---|---|---|---|---|
| 0.1 | 3D-SORT | 41.6 | 67.76 | 68.83 | 31.81 | 46.98 | 313.8 |
|  | MinkSORT | 42.45 | 68.43 | 69.52 | 33.04 | 49.02*** | 262.4*** |
| 0.5 | 3D-SORT | 42.8 | 60.83 | 82.33 | 32.55 | 57.63 | 172.8 |
|  | MinkSORT | 44.77** | 61 | 82.56 | 35.55** | 58.81* | 143.0** |
| 0.9 | 3D-SORT | 35.91 | 46.71 | 88.07 | 28.53 | 47.33 | 112.4 |
|  | MinkSORT | 37.04 | 46.73 | 88.11 | 30.3 | 48.27* | 88.4** |

## 4. Discussion

Creating an accurate representation of agro-food environments that robots can use to perform tasks like harvesting or plant maintenance accurately is challenging due to occlusions. These lead to errors as discussed previously. A common method to build world models is a tracking algorithm (Elfring et al., 2013; Persson et al., 2020; Rapado-Rincón et al., 2023). Although MOT algorithms have shown good performance in monitoring and yield estimation tasks in agro-food environments (Halstead et al., 2018, 2021; Kirk et al., 2021; Villacrés et al., 2023), counting error metrics only tell one part of the story. ID switching errors can occur even with low counting error, leading to poor performance of the robot's representation when deployed (Rapado-Rincón et al., 2023). To get a more complete picture of MOT algorithm performance, tracking metrics that consider errors missed by counting metrics are necessary. Better tracking performance leads to a better representation.

MinkSORT, achieves a HOTA of 44.77% and a MOTA of 58.81%, given an object detection recall of 61% and precision of 82.56%. Villacrés et al. (2023) achieved a MOTA between 40.5% and 69.3% with deepSORT tracking apples in images, given a detection recall of 41.1% and 70.5% and precision of 99.1% and 98.8%, respectively. They obtained these results using video sequences where individual frame distances are much smaller than in the dataset used in this work. MinkSORT achieves similar performance with fewer frames and larger distances between frames by tracking objects using 3D information.

The small training dataset that was used, which contains only 200 images from two plants and a total of 58 different tomatoes, is much smaller than existing implementations of 2D deepSORT in agro-food environments. Kirk et al. (2021) used 772 different strawberries to train their feature extractor network, while Villacrés et al. (2023) used approximately 1500 apples. The difference in MOTA between SORT and deepSORT is larger in Villacrés et al. (2023) than in the presented results between 3D-SORT and MinkSORT. This difference could be due to the difference in training data. Deep neural networks improve performance with more data, suggesting that MinkSORT could benefit greatly from additional data. This idea is further supported by the training values presented in Figure 5. A large difference can be found between the loss and top 1 accuracy of train and validation sets, suggesting a partial overfitting on the small training set. More training data could solve this problem.

Table 2 shows that the used detection algorithm can be improved further. Two-step tracking algorithms like ours are highly susceptible to the detection step's performance (Bewley et al., 2016). Villacrés et al. (2023) performed a sensitivity analysis of the detection performance's effect by using ground truth bounding boxes as input instead of an object detector and randomly removing several percentages of these boxes to emulate different recall performances. They found that the difference between SORT and deepSORT becomes smaller when the detection algorithm's performance increases. Unfortunately, a similar experiment with MinkSORT could not be performed because the ground truth data lacked the pixel-level segmentation per object required by the detection method. In the detection confidence experiments, it was proved that the differences between 3D-SORT and MinkSORT remain consistent over a small range of precision and recall detection values.

When $\lambda$ is set to 1 and only the features generated by the feature extractor are used for data association, the performance is similar regardless of the sequence order. This suggests that MinkSORT can extract valuable information from the noisy point clouds of the individual tomatoes. However, the overall performance of MinkSORT decreases when the sequence order is random. This could be due to the limited context about the scene that MinkSORT can access. For instance, relations between objects in the scene are not considered. End-to-end deep learning approaches like Trackformer (Meinhardt et al., 2022) have the potential to learn such relations since they process the whole image with all its objects at once.

# 5. Conclusions and future work

In this work, MinkSORT was developed, a 3D tracking algorithm that successfully generates tracking features from point clouds of tomatoes using a neural network. It was demonstrated how MinkSORT can improve the tracking accuracy of a tracking algorithm based on the Kalman filter and the Hungarian algorithm when used to track tomatoes in real-world greenhouse tomato plants. This improvement can be of up to 1.97 points when frames are processed in sequential order and up to 7.72 points in more challenging situations when frames are processed in random order. It was also showed how the use of contrastive loss functions can further improve the performance of MinkSORT up to 2.91 points in random order sequences. Additionally, an evaluation of the effect of using detection algorithms with different recall and precision values was conducted, which showed that MinkSORT improves the performance regardless of the recall and precision values of the detection system.

For future research, it is planed to compare MinkSORT to an end-to-end deep learning detection and tracking algorithm (Meinhardt et al., 2022). To ensure a fairer comparison and more robust performance, it is intended to collect more data under different environmental and lighting conditions. This will enable us to extend the comparison to different environmental characteristics.

## CRediT author statement

David Rapado-Rincón: Conceptualisation, Methodology, Software, Investigation, Data Processing, Writing - Original draft; Eldert J. van Henten: Conceptualisation, Writing - Review & Editing, Supervision, Funding acquisition; Gert Kootstra: Conceptualisation, Writing - Review & Editing, Supervision, Funding acquisition.

## Funding

This research is part of the project Cognitive Robotics for Flexible Agro-food Technology (FlexCRAFT), funded by the Netherlands Organisation for Scientific Research (NWO) grant P17-01.

## Declaration of competing interest

It is declared that there are no personal and/or financial relationships that have inappropriately affected or influenced the work presented in this paper.

26